\documentclass{article}

\usepackage{arxiv}

\usepackage[utf8]{inputenc} 
\usepackage[T1]{fontenc}    
\usepackage{hyperref}       
\usepackage{url}            
\usepackage{booktabs}       
\usepackage{amsfonts}       
\usepackage{nicefrac}       
\usepackage{microtype}      
\usepackage{lipsum}
\usepackage{graphicx}
\usepackage{amsmath}
\graphicspath{ {./images/} }

\title{Autoregressive deep learning for real-time simulation of soft tissue dynamics during virtual neurosurgery}

\author{
 Fabian Greifeneder \\
  Research Unit Medical Informatics\\
  RISC Software GmbH, Hagenberg, Austria \\
   \And
  Wolfgang Fenz \\
  Research Unit Medical Informatics\\
  RISC Software GmbH, Hagenberg, Austria \\
  \And
  Benedikt Alkin \\
  Emmi AI GmbH, Linz Austria \\
  ELLIS Unit Linz, Institute for Machine Learning, JKU Linz, Austria
\And
  Johannes Brandstetter \\
  Emmi AI GmbH, Linz Austria \\
  ELLIS Unit Linz, Institute for Machine Learning, JKU Linz, Austria  \And
  Michael Giretzlehner \\
  Research Unit Medical Informatics\\
  RISC Software GmbH, Hagenberg, Austria \\
  \And
  Philipp Moser \\
  Research Unit Medical Informatics\\
  RISC Software GmbH, Hagenberg, Austria \\
  philipp.moser@risc-software.at
}

\begin{document}
\maketitle
\begin{abstract}
Accurate simulation of brain deformation is a key component for developing realistic, interactive neurosurgical simulators, as complex nonlinear deformations must be captured to ensure realistic tool–tissue interactions. However, traditional numerical solvers often fall short in meeting real-time performance requirements. To overcome this, we introduce a deep learning-based surrogate model that efficiently simulates transient brain deformation caused by continuous interactions between surgical instruments and the virtual brain geometry. Building on Universal Physics Transformers, our approach operates directly on large-scale mesh data and is trained on an extensive dataset generated from nonlinear finite element simulations, covering a broad spectrum of temporal instrument-tissue interaction scenarios. To reduce the accumulation of errors in autoregressive inference, we propose a stochastic teacher forcing strategy applied during model training. Specifically, training consists of short stochastic rollouts in which the proportion of ground truth inputs is gradually decreased in favor of model-generated predictions. Our results show that the proposed surrogate model achieves accurate and efficient predictions across a range of transient brain deformation scenarios, scaling to meshes with up to 150,000 nodes. The introduced stochastic teacher forcing technique substantially improves long-term rollout stability, reducing the maximum prediction error from 6.7~mm to 3.5~mm. We further integrate the trained surrogate model into an interactive neurosurgical simulation environment, achieving runtimes below 10~ms per simulation step on consumer-grade inference hardware. Our proposed deep learning framework enables rapid, smooth and accurate biomechanical simulations of dynamic brain tissue deformation, laying the foundation for realistic surgical training environments.
\end{abstract}


\section{Introduction}
\label{sec:introduction}

Surgical simulators are designed to offer safe, controlled, and anatomically realistic environments for surgical training, preoperative planning, and intraoperative guidance. These systems have demonstrated significant potential in enhancing both the technical skills and clinical reasoning of medical trainees, while also addressing key challenges such as variability in clinical training experiences and concerns about patient safety~\cite{Zhang2024, Larsen2012}. Surgical simulators are applied across a broad spectrum of clinical procedures, supporting tasks from endoscopic sinus surgery to hip arthroscopy~\cite{Varshney2014, Bauer2019}. In high-risk interventions (particularly those involving potential neurological damage or hemorrhage) virtual training environments are especially valuable, for example in neurosurgical contexts such as brain tumor resections and aneurysm clipping interventions~\cite{Azarnoush2015, gmeiner2018, Joseph2020}.

A core requirement for neurosurgical simulators is the realistic simulation of soft tissue deformation~\cite{meier_real-time_2005}, which is essential for accurately capturing the biomechanical behavior of soft tissues under surgical manipulation~\cite{Zhang2018}. These deformation processes are typically modeled using nonlinear elasticity theory, resulting in systems of partial differential equations (PDEs) that describe the tissue's dynamic response to applied forces. Achieving fast and high-fidelity solutions to these equations remains a significant challenge~\cite{Nguyen2020}. Real-time interactivity, however, is crucial for delivering immediate visual feedback, enabling haptic responses, and allowing dynamic adjustments to surgical plans based on user-controlled tool-tissue interactions. 

Classical methods for solving systems of PDEs commonly rely on the finite element method (FEM)~\cite{brenner_mathematical_2008}. Achieving physiologically accurate simulations often requires nonlinear material formulations and the conversion of patient-specific geometries into high-resolution meshes, both of which demand computationally expensive iterative solvers. Moreover, the brain's biomechanical properties (e.g., viscoelasticity, heterogeneity, and sensitivity to small-scale forces) make high-fidelity modeling particularly challenging. These factors necessitate high spatial and temporal resolution to realistically simulate responses to surgical actions. The complexity is further compounded by the introduction of surgical tool interactions, which constitute dynamic boundary conditions and contact mechanics~\cite{Misra2008}. These factors collectively result in significant computational overhead. Although explicit methods like Total Lagrange Explicit Dynamics (TLED)~\cite{miller_total_2007} offer performance benefits in certain scenarios, a trade-off exists between precision and computational efficiency, particularly for highly detailed simulations of complex tissue interactions. As a result, conventional numerical methods can be impractical for real-time applications in complex neurosurgical simulators, highlighting the need for alternative approaches.

In recent years, deep neural network (DNN)-based surrogate models have emerged as computationally efficient complements or alternatives to traditional numerical PDE solvers. Such surrogate models have demonstrated significant potential in capturing complex physical phenomena across various domains of science and engineering~\cite{Zhang2023, Thuerey2021}. Notable applications include fluid dynamics~\cite{Raissi2019, Vinuesa2022}, solid mechanics~\cite{Deshpande2023}, and weather forecasting~\cite{Bodnar2024, bi2022, lam2023}. Once trained on data pre-computed with numerical solvers, these surrogate models allow for highly accurate and fast predictions of simulation outcomes. These benefits make surrogate models an ideal candidate to be be integrated into surgical simulation pipelines, replacing the computationally intensive numerical solver components.

Common surrogate models often build upon convolution-based architectures such as U-Nets~\cite{Gupta2022} or ResNets~\cite{gregory_equivariant_2024}. These methods typically operate on regular grids and require projecting mesh-based data onto grids~\cite{Mendizabal2020}. This projection step can lead to a loss of geometric detail, and imposes additional memory and computational burdens when dealing with complex geometries. As a result, convolution-based architectures can be cumbersome and inefficient when working directly with mesh-based simulation data.

Many recent surrogate models rely on neural operators~\cite{li_neural_2020,Kovachki2021,Alkin2024,li_fourier_2021,Wang2024,Li2023}, which are a class of machine learning models designed to learn mappings between function spaces, such as those arising in PDEs. Instead of approximating specific solutions at fixed input points, neural operators aim to learn the entire solution operator, i.e., a function that maps input functions (e.g., initial or boundary conditions) to output functions (e.g., solution field). Such capabilities make neural operators particularly well-suited for capturing the fine-grained, spatially continuous deformations.

We chose Universal Physics Transformers (UPT)~\cite{Alkin2024} as a backbone architecture due to their flexible encoder-approximator-decoder blocks and demonstrated applicability to multi-physics~\cite{alkin_neuraldem_2025} and computational fluid~\cite{alkin_ab-upt_2025} dynamics. While other backbones such as OFormer~\cite{li_transformer_2023}, Transolver~\cite{wu_transolver_2024}, MeshGraphNet~\cite{pfaff_learning_2021}, or DoMINO~\cite{ranade_domino_2025} could in principle be used, recent results~\cite{alkin_ab-upt_2025} suggest that UPT offer superior performance and scalability compared to these alternatives.

In this work, we develop a UPT-based surrogate model, tailored for predicting transient soft tissue deformation caused by continuous interactions of surgical instruments with the virtual brain mesh. While there has already been prior work on deep learning-accelerated simulation of soft tissue deformation~\cite{shahbazi_neural-augmented_2025,salehi_physgnn_2022,liu_real-time_2020,karami_real-time_2023,zhang_neural_2019}, these reports often rely on convolutional, cellular, or graph-based architectures, which can be computationally demanding and less scalable when applied to high-resolution meshes. To the best of our knowledge, no prior work has demonstrated autoregressive inference, which is a critical component for modeling transient tissue dynamics enabling a tightly coupled interaction loop between the surgical instrument and the virtual brain mesh. Our high-fidelity approach constitutes a promising path for future neurosurgical simulations, providing surgeons with a realistic training environment to practice and refine complex surgical interventions.

\section{Methods}
\label{sec:methods}



\subsection{Overview of the autoregressive framework} 
\begin{figure}[t]
\hspace{-6pt}\includegraphics[width=1.0\textwidth]{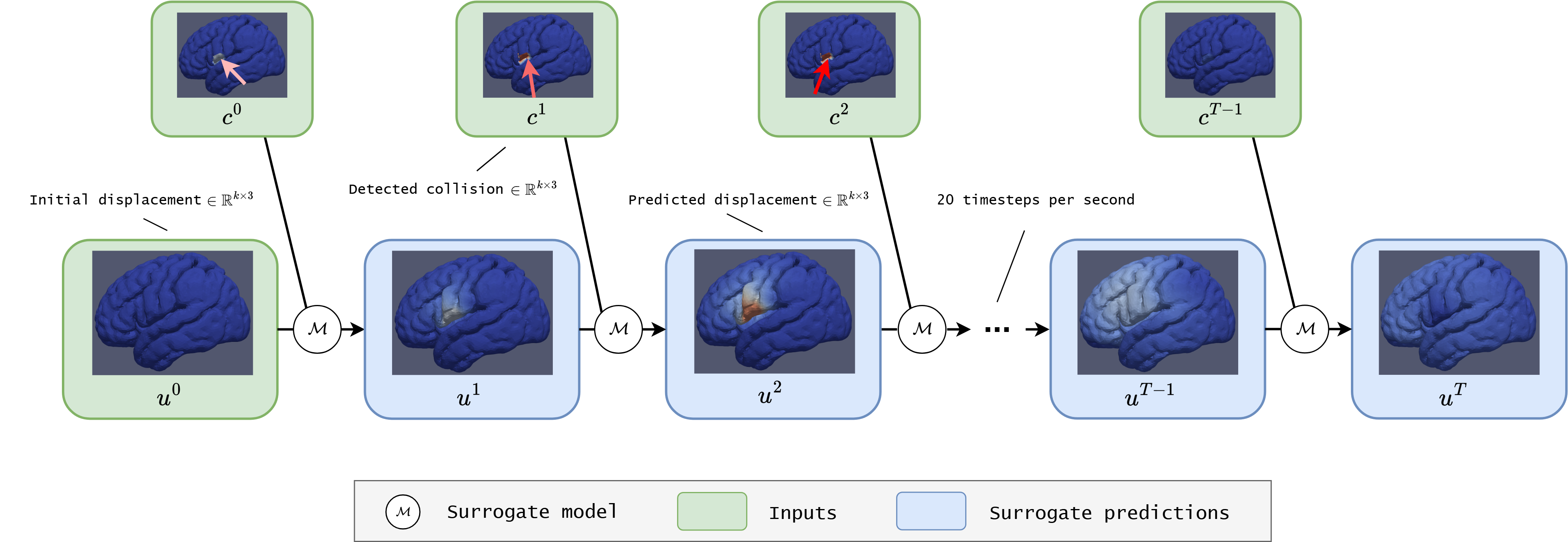}
\caption{Overview of our framework for autoregressive prediction of brain tissue deformation. At each time step, the model receives the previously predicted displacement and the current collision field as input. The overlaid arrows on the collision fields indicate the surgical instrument’s pushing direction.}
\label{fig:framework_overview}
\end{figure}

At the core of our methodology is an autoregressive prediction framework (Figure~\ref{fig:framework_overview}), in which the deep learning-based surrogate model iteratively predicts brain deformation, with each step taking the output of the previous prediction as input. These predictions depend on dynamic interactions between a virtual surgical instrument and brain tissue. As the surgeon manipulates the instrument within the simulator (applying varying forces, changing directions, or engaging with different anatomical regions) contact points and interaction forces evolve continuously. 

An initial three-dimensional displacement field \( u^0~\in~\mathbb{R}^{K \times 3} \) is given at time step \( t=0 \) (typically the rest position, i.e., the undeformed brain mesh), where $K$ denotes the number of mesh nodes. Given a temporal sequence of positions corresponding to the motion of a pushing or releasing surgical instrument, our goal is to stepwise predict the resulting brain deformation in an autoregressive fashion: \begin{equation}
(u^t, c^t) \xrightarrow{\mathcal{M}} u^{t+\Delta t}.
\end{equation}
Specifically, we want to learn a function, i.e., a neural surrogate model \( \mathcal{M} \), that maps the current displacement field \( u^t \) to the subsequent one \( u^{t+\Delta t} \), depending on the current collision state \( c^t \) between the instrument and the brain model. 

\subsection{Surrogate model and training strategy}
 %

\subsubsection{Deep learning architecture} 
Our surrogate model builds on the Universal Physics Transformer (UPT) framework~\cite{Alkin2024}, a highly efficient and scalable paradigm for learning physical dynamics. Here, the model $\mathcal{M}$ encodes the displacement field \( u^t \in \mathbb{R}^{K \times 3} \) at $K$ mesh nodes at time \( t \) into a compact latent representation \( z^t \in \mathbb{R}^{d_\text{latent}} \). This latent signal is propagated forward in time by a stack of transformer blocks and subsequently decoded back onto the full brain mesh to predict the displacement field \( u^{t + \Delta t} \in \mathbb{R}^{K \times 3} \) at the next time step.

\begin{figure}[t]
\centerline{\includegraphics[width=1\textwidth]{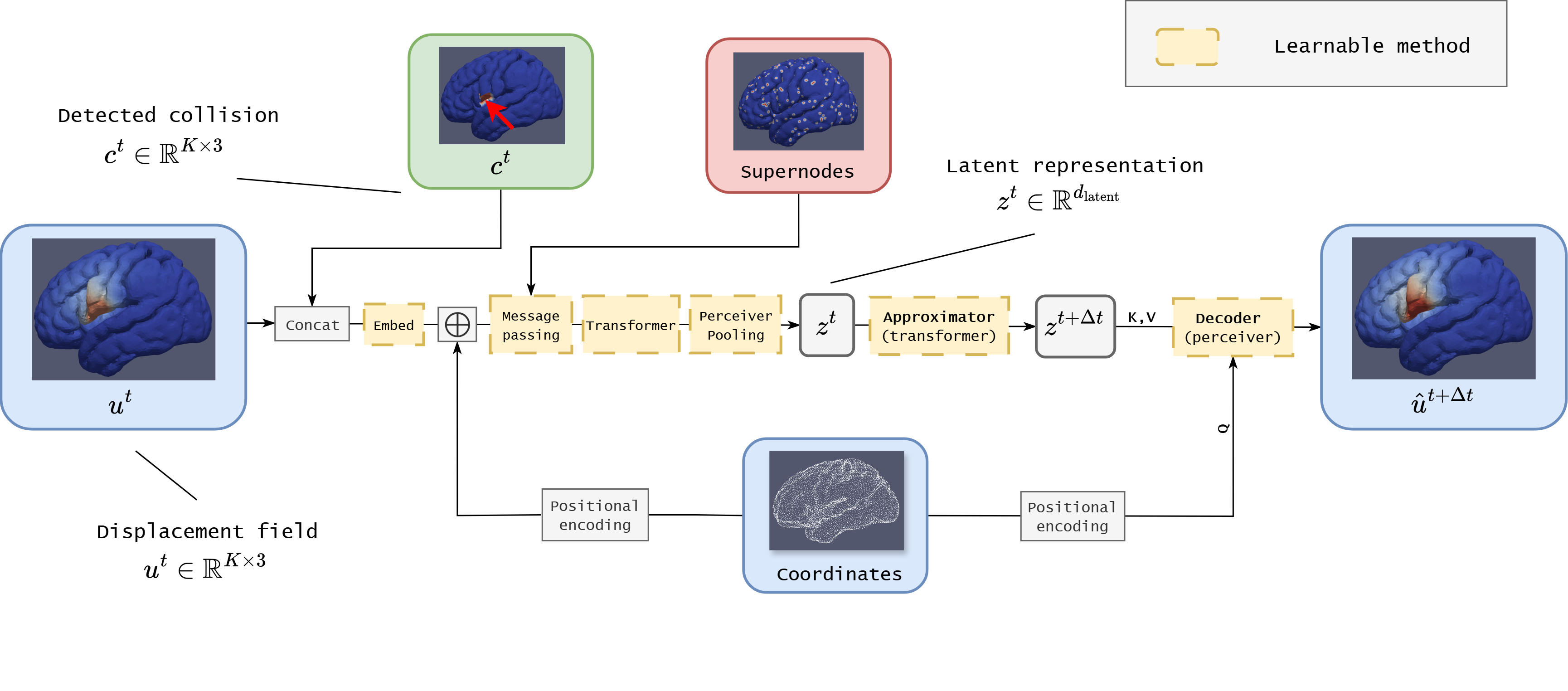}} \vspace{-10pt}
\caption{Prediction of the displacement field at the next time step. The current displacement field is concatenated with the respective collision field. Spatial context is incorporated through positional encodings after passing through a learned embedding layer. The input is then mapped into latent space via a message passing layer, that aggregates information at randomly selected supernodes. This latent signal is propagated forward through a stack of transformer blocks and evaluated at the mesh coordinates using perceiver-style cross-attention.}
\label{fig:upt_architecture}
\end{figure}

To preserve spatial context, the collision field $c^t \in \mathbb{R}^{K \times 3}$ is defined on all mesh nodes of the brain model but contains nonzero values only at nodes in contact with the surgical instrument, where it encodes the displacement induced by the current instrument position. As illustrated in Figure~\ref{fig:upt_architecture}, the displacement field \( u^t \) is concatenated with the corresponding collision field \( c^t \) in the \textit{encoder} $\mathcal{E}$, resulting in a combined tensor of shape \( K \times 6 \), which is then projected into a hidden space of dimension \( K \times d_\text{embed} \). Spatial information is preserved employing positional encoding on the mesh coordinates~\cite{Vaswani2017}. This resulting representation is processed by a message passing (MP) layer~\cite{Bronstein2021}, which aggregates local features at a set of \( n_S \ll K \) randomly sampled \textit{supernodes}. These supernode features are then processed by transformer blocks~\cite{Vaswani2017}, and subsequently mapped into a latent space $\mathbb{R}^{d_\text{latent}}$ via perceiver-style iterative cross-attention~\cite{Jaegle2021} with a learnable latent array. The latent representation \( z^t \) is then compact enough to be propagated efficiently through the temporal transformer stack, referred to as the \textit{approximator} \(\mathcal{A}\). Finally, a perceiver-style \textit{decoder} \(\mathcal{D}\) maps the latent state \( z^{t+\Delta t} \) back to the original mesh using cross-attention, where positional encodings of the mesh nodes serve as queries and the latent features provide keys and values.

In summary, the surrogate model $\mathcal{M}$ predicts the next displacement field \( \hat{u}^{t+\Delta t} \) from the current state \( u^t \) via the following sequence of operations:
$$\mathcal{M}: u^t \in \mathbb{R}^{K\times 3} \xrightarrow{\text{ concatenate with } c^t \text{ }} \mathbb{R}^{K\times 6} \xrightarrow{\text{ embed }} \mathbb{R}^{K\times d_\text{embed }} $$\begin{equation} \xrightarrow{\text{ MP }} \mathbb{R}^{n_S\times d_\text{embed}} \xrightarrow{\text{ transformer }} \mathbb{R}^{n_S\times d_\text{embed}} \xrightarrow{\text{ perceiver }} z^{t} \in \mathbb{R}^{d_\text{latent}} \end{equation}$$ \xrightarrow{ \text{ }\mathcal{A} \text{ }} z^{t+\Delta t} \in \mathbb{R}^{d_\text{latent}} \xrightarrow{\text{ }\mathcal{D} \text{ }} \hat{u}^{t+\Delta t} \in \mathbb{R}^{K\times 3}.$$

\subsubsection{Stochastic teacher forcing} \label{sec:tf_vs_auto}

While numerical ground truth displacement fields are available during training, only the brain’s initial rest position is known at inference time. All subsequent deformations must then be predicted purely in an autoregressive manner, guided by the surgeon’s instrument movements. To account for this discrepancy, we distinguish between two types of predictions depending on the model's input: \textit{teacher-forced} and \textit{autoregressive}. In teacher-forced predictions, the model receives the numerically precomputed ground truth displacement field \( u^t \) as input and predicts the displacement at the next time step \( \hat{u}^{t+\Delta t} \). In contrast, \textit{autoregressive} predictions simulate forward dynamics by recursively feeding the model's previous outputs as subsequent inputs. Starting from the ground truth displacement \( u^{t_0} \), the model is applied repeatedly, using each predicted displacement as input for the next step:
\begin{equation} 
u^{t_0} \xrightarrow{\mathcal{M}} \hat{u}^{{t_1}} \xrightarrow{\mathcal{M}} \hat{u}^{t_2} \xrightarrow{\mathcal{M}} \ldots \xrightarrow{\mathcal{M}} \hat{u}^{t_{ S_\text{Auto}}},
\end{equation}
where $t_i=t_{i-1}+\Delta t$ and \( S_\text{Auto} \geq 2 \) denotes the number of autoregressive steps.

Relying exclusively on teacher-forced training (i.e., using precomputed ground truth displacements as inputs) would result in a distribution shift at inference~\cite{brandstetter2023, bengio2015}. The model would learn to operate under ideal conditions with perfect inputs, but would be expected to perform under autoregressive conditions during deployment, leading to compounding errors. Conversely, training the model with long autoregressive sequences from the outset could destabilize learning, as it would be exposed to highly inaccurate predictions as inputs early in training.

\begin{figure}
\centerline{\includegraphics[width=0.68\columnwidth]{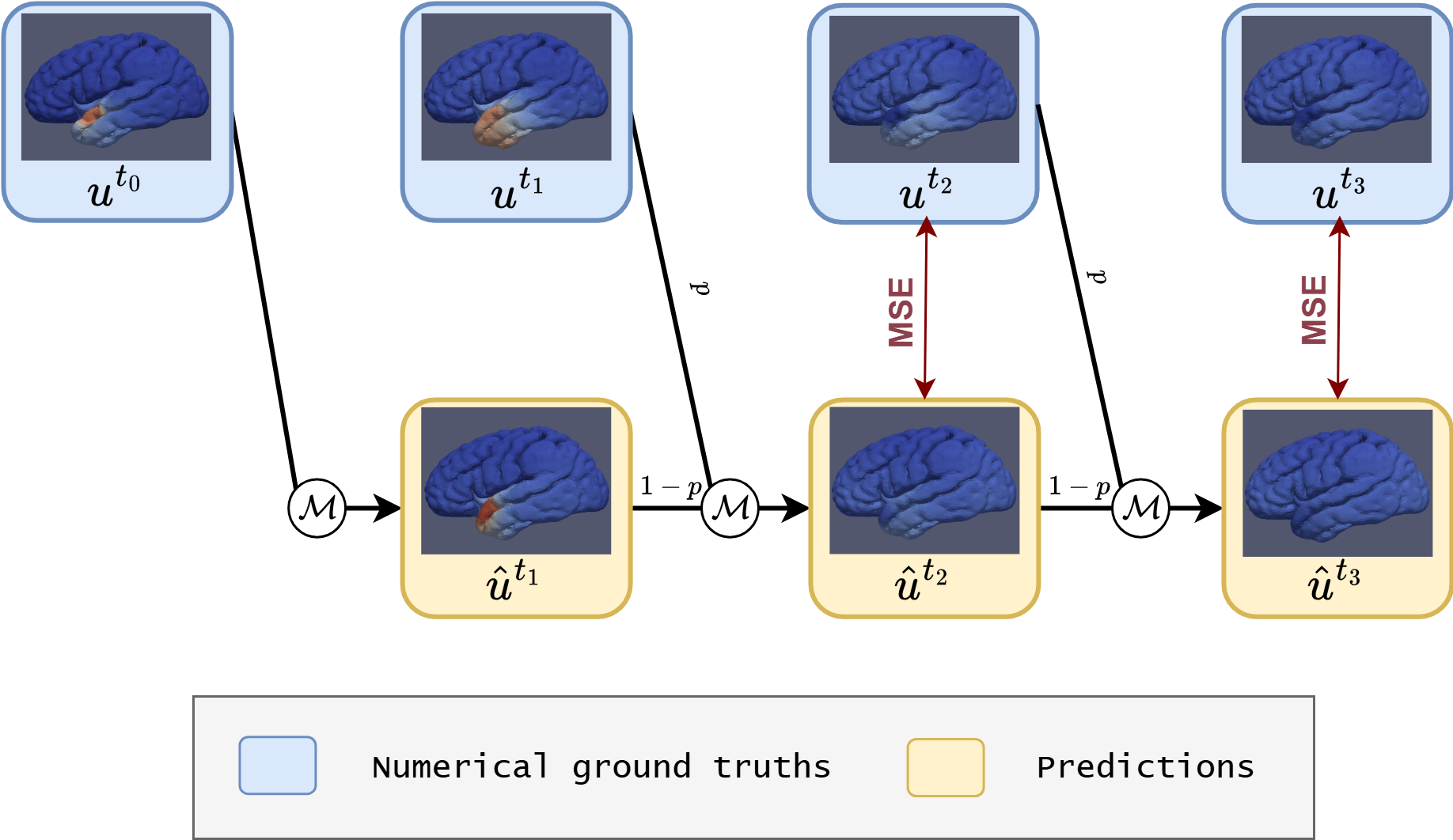}}
\caption{\textit{Stochastic teacher forcing} step for a rollout with window size $S_{\text{STF}}=3$. At each prediction step the input is sampled as either the numerically precomputed ground truth $u^{t_i}$ (teacher-forced) or the predicted value $\hat{u}^{t_i}$ (autoregressive), according to the probability $p$. The first prediction is excluded for the loss computation. Collision fields are omitted for clarity.}
\label{fig:stochastic_teacher_forcing}
\end{figure}

Hence, to balance training and inference stability, we introduce \textit{stochastic teacher forcing}, which combines the benefits of both teacher-forced and autoregressive training strategies. Similar to fully autoregressive rollouts, the model is applied repeatedly over a rollout window, i.e., a subsequence, of $S_{\text{STF}} \geq 2$ successive time steps. As visualized for $S_{\text{STF}} = 3$ in Figure~\ref{fig:stochastic_teacher_forcing}, at each time step $t_i$, the input $x^{t_i}$ is selected according to a teacher forcing probability $p \in [0,1]$: with probability $p$, the model receives the ground truth displacement $u^{t_i}$; with probability $1 - p$, it uses its own previous prediction $\hat{u}^{t_i}$. Setting the teacher forcing probability to $p = 0$ yields a fully autoregressive regime, while using $p = 1$ and a rollout window size of $S_{\text{STF}} = 2$ recovers the teacher-forced setup.

Although any value of $p \in [0,1]$ is theoretically valid, we found it beneficial to start training with a teacher forcing probability close to $1$ and gradually decrease it throughout the training process. This scheduling strategy stabilizes learning in the early phases while progressively exposing the model to autoregressive conditions, helping it adapt effectively for fully autoregressive rollouts at inference time.

During training, gradients can either be propagated through the entire sequence of autoregressive steps or restricted to the current prediction only. Backpropagating through time enables better temporal credit assignment and improved long-term consistency but may increase the risk of exploding or vanishing gradients. Blocking gradients avoids this risk but can limit the model's ability to correct early-step errors. In our case, we chose to backpropagate through time and found the risk to be manageable in practice, as gradient norms remained stable throughout training. Importantly, due to the stochastic teacher forcing schedule, gradients are initially blocked more often (when teacher-forced inputs dominate) and are propagated through longer sequences as training progresses, since teacher-forced inputs inherently act as gradient blockers.

\subsection{Data generation}
We performed extensive finite element simulations to generate a large dataset of realistic brain deformations resulting from transient interactions between surgical instruments and brain tissue. This data generation process captured key aspects of instrument-tissue interaction by randomly sampling varying contact locations, pushing directions, and displacement profiles over time, while calculating the corresponding biomechanical tissue response.

\subsubsection{Brain model and collision site selection}
\begin{figure}
\centerline{\includegraphics[width=0.48\columnwidth]{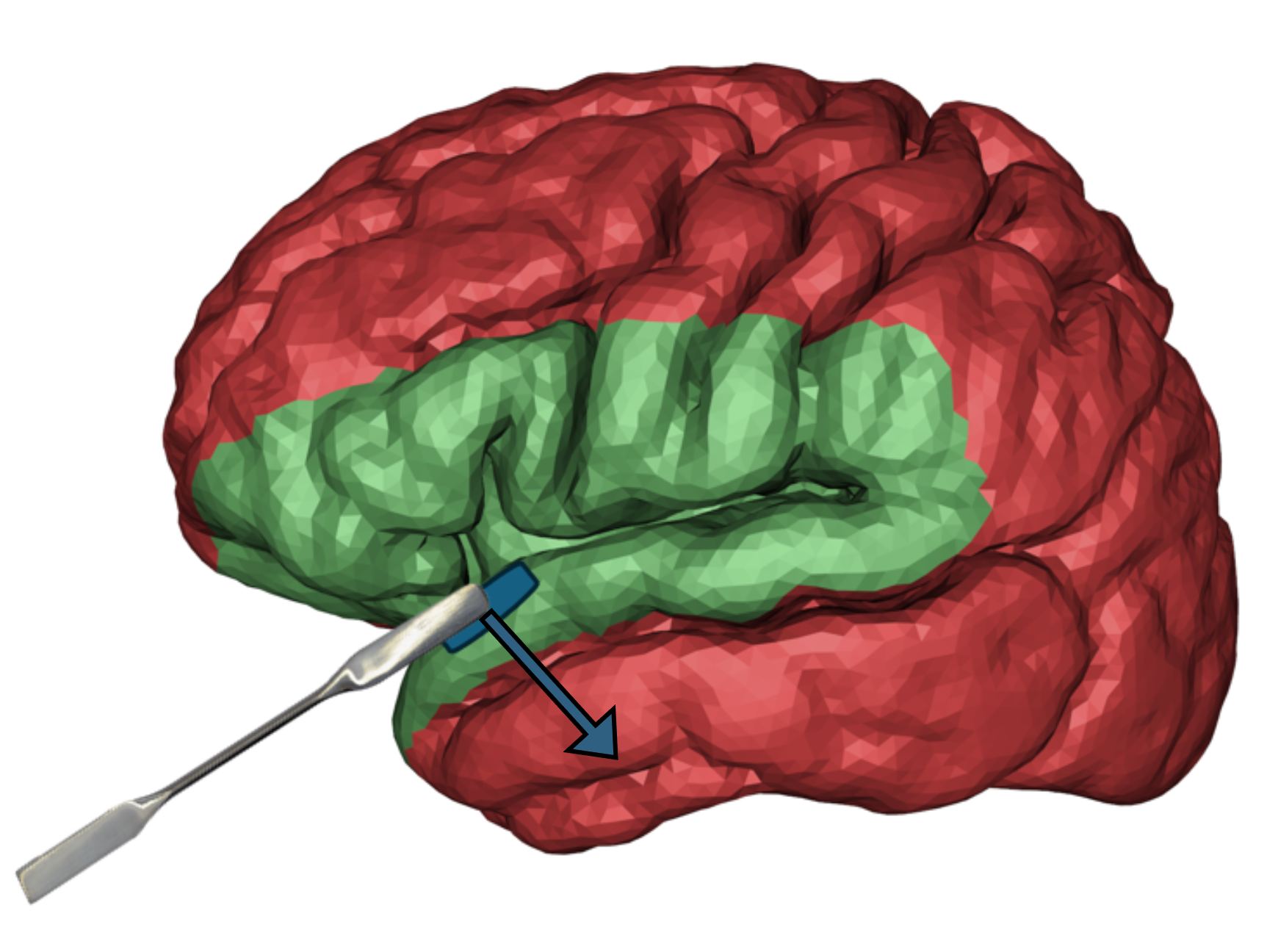}}
\caption{Visualization of the predefined collision domain (green) around the Sylvian fissure, with an example of a sampled collision site and direction (blue) and a superimposed surgical instrument (spatula) demonstrating a typical interaction scenario.}
\label{fig:collision_domain}
\end{figure}

The brain model used throughout this study was derived from the Allen Human-Brain Reference Atlas~\cite{Ding2016} and is shown in Figure~\ref{fig:collision_domain}. It consisted of $K = 21{,}670$ mesh nodes, providing a detailed representation of the brain anatomy.

We defined a collision domain, i.e., a specific region of the brain where various instrument-tissue interactions may occur: As shown in green in Figure~\ref{fig:collision_domain}, this domain was centered around the Sylvian fissure, a critical anatomical landmark in aneurysm clipping procedures where brain tissue is carefully displaced or separated during surgery.

A deformation sequence was initiated by randomly selecting a single triangular mesh face (with an edge length of 2.5 mm) from the domain. From this seed, the collision site was expanded iteratively, face by face, until it spanned up to 100 mesh faces. The resulting site was centered around the initial face and remained entirely within the predefined collision domain. An example of such a collision site is illustrated in blue in Figure~\ref{fig:collision_domain}.

\subsubsection{Transient deformation and simulation schemes}
\begin{figure}
\vspace{-10pt}
\centerline{\includegraphics[width=0.48\textwidth]{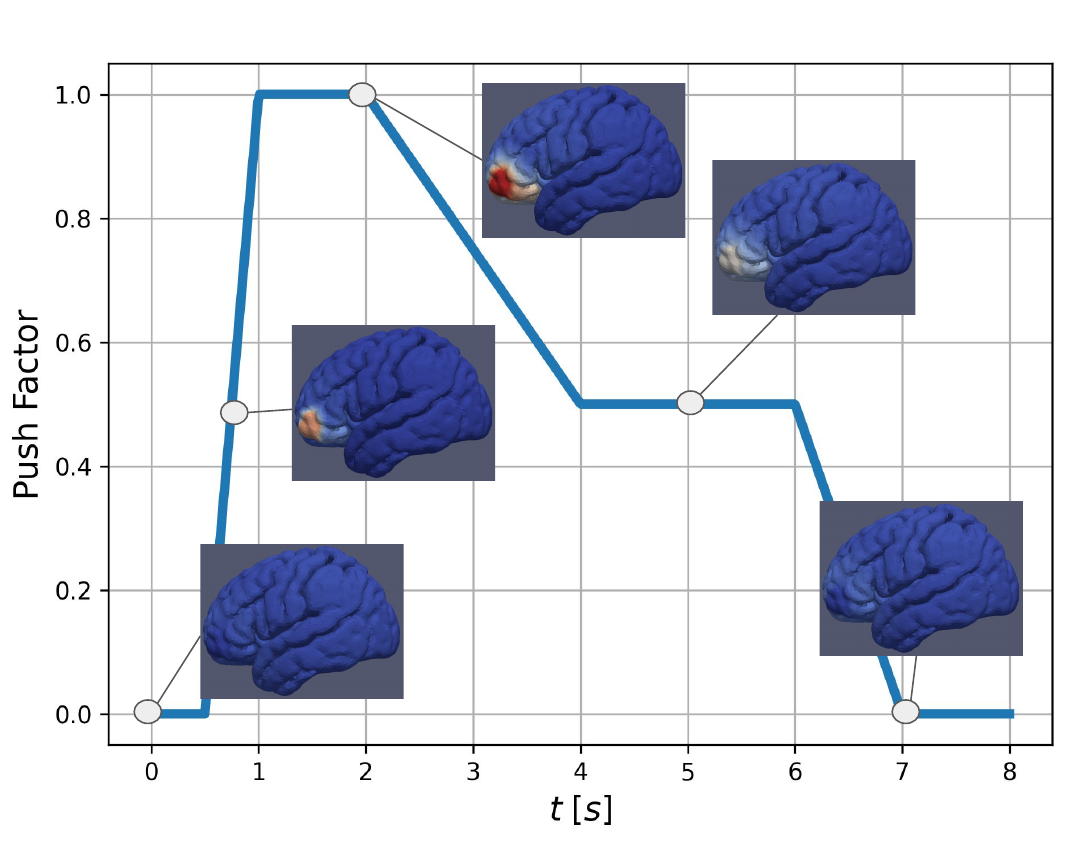}}
\caption{Example trajectory of a time-dependent push factor, with selected displacement fields overlaid at corresponding time steps to illustrate the resulting deformation progression.}
\label{fig:push_factor_trajectory}
\end{figure}

The pressing, holding, and releasing phases of the instrument-tissue interaction were simulated by scaling the normalized collision vector with a time-dependent push factor and a constant magnitude factor. This produced realistic deformation scenarios, with maximum displacements reaching up to 15~mm. Figure~\ref{fig:push_factor_trajectory} illustrates the effect of the push factor on deformation, overlaying an example push-factor trajectory with the corresponding sequence of displacement fields at certain points in time. 

The initial direction of instrument-tissue interaction was defined by the normalized surface normal vector of the seed triangle. To enhance the realism of instrument behavior, we introduced angular variability by gradually deviating this normalized collision vector using a time-dependent angle~\( \phi \), and then rotating it around its original axis by a second time-dependent angle~\( \theta \). Both angle trajectories were modeled as continuous functions of time, ensuring smooth and natural directional changes.

\begin{figure}
\vspace{-10pt}
\centerline{\includegraphics[width=0.99\textwidth]{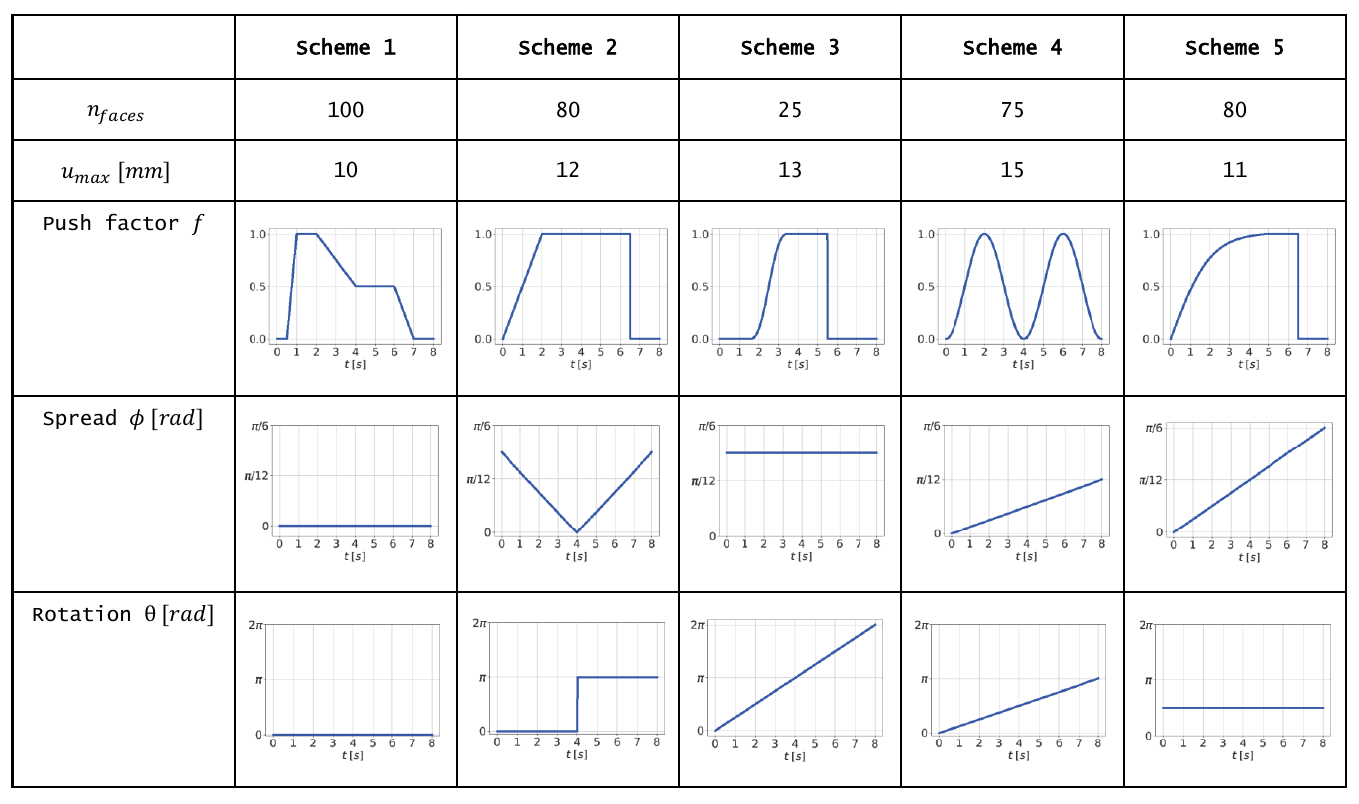}}
\caption{Simulation schemes used for data generation.}
\label{fig:simulation_schemes}
\end{figure}

For our training dataset, we defined five distinct deformation scenarios by combining different pairs of angle and push-factor trajectories. Each of these five \textit{simulation schemes}) was further characterized by a fixed maximum displacement between 10 and 15~mm and a fixed number of collision faces ranging from 25 to 100. The corresponding hyperparameters for each scheme are summarized in Figure~\ref{fig:simulation_schemes}. Both the angle and push-factor trajectories were continuous, time-dependent functions defined over the full simulation interval~\( [0, T] \). 

As an example, simulation scheme 3 generated pushing scenarios in which the instrument's push vector was offset by an angle \(\phi = \frac{\pi}{8}\) from the normal vector of the collision surface, and was continuously rotated around this normal vector to complete a full circular sweep. The push factor gradually increased until a maximum displacement of 13~mm was reached. At $t=6.5\,\mathrm{s}$ the instrument was released from the brain tissue, meaning the boundary condition enforcing fixed displacement at the collision site was lifted—visualized by an abrupt drop of the push factor to zero. While these hyperparameters remained consistent across all simulations within simulation scheme 3, the collision site was randomly sampled for each individual simulation, introducing sufficient variability.

\subsubsection{Numerical simulation} \label{sec:dataset}
At each time step, the displacements at the mesh nodes belonging to the collision site were set to the current collision vector, representing the first boundary condition. As a second boundary condition, nodes at the corpus callosum, which structurally connects the two brain hemispheres, were kept fixed. Given these boundary conditions, the deformation of the remaining mesh nodes was computed numerically using a nonlinear Neo-Hookean elasticity model with an explicit finite-element algorithm (TLED~\cite{miller_total_2007}) implemented in our in-house simulation framework~\cite{Fenz2016}. Tissue properties were set according to established values in the literature, with a Young’s modulus of 3000 Pa and a Poisson’s ratio of 0.49~\cite{gmeiner2018}. The time step used for the simulations was 0.1~ms, displacement fields were output every 500 steps, thus resulting in an effective time step of $\Delta t = 50$~ms.

For each simulation scheme in Figure~\ref{fig:final_rollout_cross}, we generated 210 simulations (totaling, 1050 simulations), each spanning \( T = 8\,\mathrm{s}\) at a temporal resolution of 20 steps per second. This resulted in deformation trajectories consisting of 160 displacement fields per simulation. At each time step \( t \), the displacement field \( u^t \in \mathbb{R}^{K \times 3} \) contained displacement vectors for all \( K = 21{,}679 \) mesh nodes. The numerical computation of the next displacement \( u^t \to u^{t+\Delta t} \), with \( \Delta t = 50\,\mathrm{ms} \), took approximately $1.68 \,\mathrm{s}$.

\section{Experimental setup}
\subsection{Model training and configuration} 
As data preparation, displacement fields were normalized independently in the \( x \)-, \( y \)-, and \( z \)-directions, with mean and standard deviation computed over the entire training set. Additionally, the mesh coordinates were scaled and shifted to lie within \([0,200]^3\) to ensure compatibility with the positional encoding mechanism, which assumes positive coordinate values~\cite{Vaswani2017}.

We first conducted a hyperparameter optimization to determine the most effective training strategy for robust autoregressive inference. The surrogate models used in these sections were trained on $N_\text{Train}=200$ simulations, and evaluated on $N_\text{Test}=10$, both from from simulation scheme 2. 

Using these optimized hyperparameters, we then trained a final surrogate model on $N_\text{Train}=800$ simulations equally sampled from simulations schemes 1-4. For in-distribution performance evaluation, a test set of $N_\text{Test} = 40$ simulations (10 simulations per scheme 1-4) was used. To evaluate cross-scheme generalization, we tested the model on $N_\text{Test} = 10$ simulations from simulation scheme 5, which was entirely excluded during training.

We used a UPT-based model comprising 3.4 million trainable parameters. Specifically, the architecture included $n_S = 2048$ supernodes, an embedding dimension of $d_\text{embed} = 128$, a latent dimension of $d_\text{latent} = 256 \times 256$, and a depth of 4 blocks for both the transformer and perceiver sub-modules. The model was trained using the stochastic teacher forcing MSE loss $\mathcal{L}_{\text{STF}}$ defined in Eq.~\ref{eq:mse_stf}, applied over rollouts of $S_\text{STF}$ time steps, where $S_\text{STF} \in \{3,4,5,6,7,8\}$. The teacher forcing probability $p$ decreased linearly over the course of training, i.e., $p = 1 - \frac{n}{N_\text{STF}}$, where $n$ denotes the current epoch.

Training was conducted on consumer-grade hardware (NVIDIA GeForce RTX 3080 Ti GPU with 12 GB of VRAM), for $N_\text{STF}~=~45$ epochs using stochastic teacher forcing. To further refine inference performance, the final model was trained for an additional $N_\text{Auto} = 15$ fully autoregressive epochs. Model parameters were optimized using the Lion optimizer~\cite{Chen2023} with an initial learning rate of \( 1.5~\times~10^{-5} \), a 10-epoch constant warm-up phase and a subsequent cosine decay schedule. To improve generalization, stochastic regularization via drop path~\cite{Huang2016} with probability $p_\text{drop}~=~0.15$ was applied across all training runs. Additionally, early stopping was employed, retaining the model with the lowest autoregressive MSE across 50 simulations from the test set. To accommodate memory constraints, gradients were accumulated across multiple steps until an effective batch size of 4 is reached. At test-time, evaluations were performed using all 160 time steps, i.e. a full-sequence autoregressive rollout.


\subsection{Evaluation metrics}

The Mean Squared Error (MSE) between the predicted displacement $\hat{u}^{t+\Delta t} := \mathcal{M}(u^t)$ and the ground truth $u^{t+\Delta t}$ is used to assess the performance of the surrogate model:

\begin{equation} 
\text{MSE}(u, \hat{u}) := \frac{1}{K} \sum_{i=1}^{K} \left\| u_i - \hat{u}_i \right\|^2,
\end{equation}  where $u_i \in \mathbb{R}^3$ denotes the displacement vector at mesh node $i$, $K$ is the total number of nodes and $||\cdot||$ denotes the Euclidean norm.

To further evaluate spatial accuracy, we computed the Hausdorff Distance between the predicted and ground truth mesh node positions. This metric was evaluated within a local radius of $r = 50$ units around the collision site to reduce computational cost. For point sets $A$ and $B$, the Hausdorff Distance is defined as:
\begin{equation} 
d_H(A, B) := \max \left\{ \sup_{a \in A} \inf_{b \in B} \|a - b\|, \; \sup_{b \in B} \inf_{a \in A} \|b - a\| \right\}.
\end{equation}

We complement our evaluation with a medically relevant measure by computing the maximum prediction error over all mesh nodes and spatial dimensions: \begin{equation} \text{Err}_\text{max}(u,\hat{u}) := \max_{i\in\{1,\ldots,K\},j\in\{1,2,3\}}|u_{i,j}-\hat{u}_{i,j}|.\end{equation} 

The MSE was employed as the default loss function $L$ in both teacher-forced and autoregressive training regimes. The teacher-forced loss at a single time step was given by:
\begin{equation}
\mathcal{L}_\text{tf}(\mathcal{M}; u^t) = L(u^{t+\Delta t}, \mathcal{M}(u^t)).
\label{eq:mse_tf}
\end{equation}
The autoregressive loss over a rollout with window size \( S_{\text{Auto}} \) was:
\begin{equation}
\mathcal{L}_{\text{Auto}}(\mathcal{M}; u^{t_1}, S_\text{Auto}) = \sum_{i=2}^{S_\text{Auto}} L\left(u^{t_i}, \mathcal{M}(\hat{u}^{t_{i-1}})\right),
\label{eq:mse_auto}
\end{equation} where the sum was scaled by $\frac{1}{S_\text{Auto}-1}$ to normalize the loss and the first prediction was excluded, as it is based on a ground truth input. If the maximum prediction error was chosen as the loss function $L$, the corresponding autoregressive loss $\mathcal{L}_\text{Auto}^\text{Max}$ was computed as the maximum error over all time steps rather than the normalized sum.

The stochastic teacher forcing loss was given by:

\begin{equation}
\mathcal{L}_{\text{STF}}(\mathcal{M}; u^{t_1}, S_\text{Auto}, p) = \sum_{i=2}^{S_\text{STF}} L\left(u^{t_i}, \mathcal{M}(\hat{x}^{t_{i-1}})\right),
\label{eq:mse_stf}
\end{equation} where the model input at each step was sampled as:

\begin{equation} 
\hat{x}^{t_i} =
\begin{cases}
\hat{u}^{t_i}, & \text{with probability } 1 - p \\
u^{t_i}, & \text{with probability } p \\
\end{cases}
\end{equation} 

As with the autoregressive loss $\mathcal{L}_{\text{Auto}}$, the first prediction was excluded from the loss computation, and the total loss is scaled by $\frac{1}{S_{\text{STF}} - 1}$ during training.

\section{Results} \label{sec:results}

\subsection{Impact of stochastic teacher forcing} \label{sec:results_stf}
\begin{figure}
\centerline{\includegraphics[width=0.48\columnwidth]{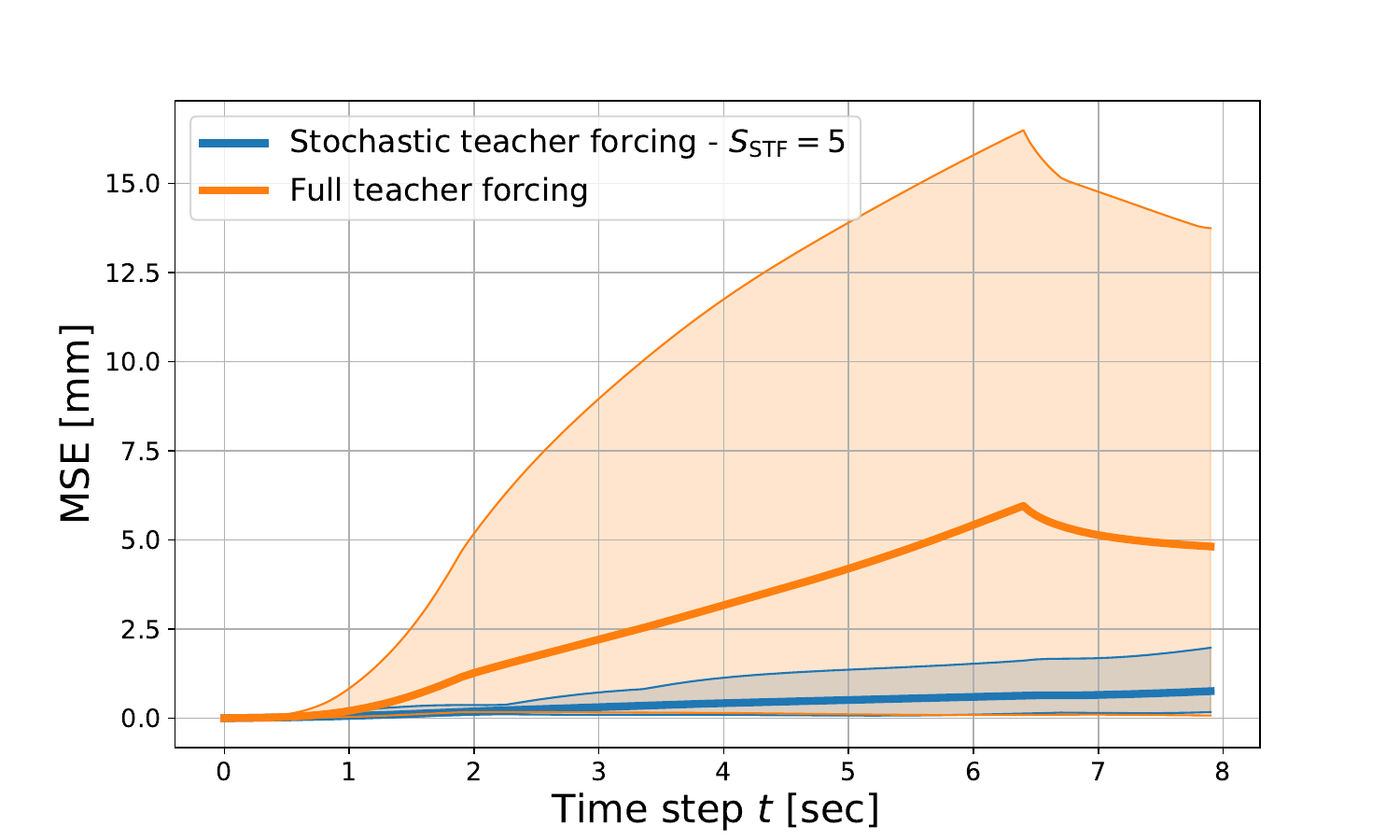}}
\caption{Full teacher forcing vs. stochastic teacher forcing: Mean autoregressive MSE over 10 test simulations (drawn from simulation scheme 2) surrounded by a shaded area bounded by the respective minimum and maximum MSE values.}
\label{fig:results_error_accumulation}
\end{figure}

To mitigate error accumulation caused by the input distribution shift between training (using ground truth inputs) and inference (using the model’s own predictions), we propose a stochastic teacher forcing strategy during training. Figure~\ref{fig:results_error_accumulation} compares MSE curves over time for 10 autoregressive rollouts of test set simulations: One model was trained solely with teacher forcing (orange), while the other was trained using stochastic teacher forcing (blue) with rollout window size $S_\text{STF} = 5$ and a linearly decaying teacher forcing probability. The purely teacher-forced model exhibited instability during longer rollouts, reflected in an autoregressive test loss of $\mathcal{L}_\text{Auto}^\text{MSE} = 1.90$, whereas the stochastically trained model, with a significantly lower test loss of $\mathcal{L}_\text{Auto}^\text{MSE} = 0.35$, adapted to the changing input distribution and maintains stability by mitigating error accumulation. Moreover, the average maximum prediction error $\mathcal{L}_\text{Auto}^\text{Max}$ was significantly reduced by $48\%$ from $6.68$~mm to $3.50$~mm using stochastic teacher forcing.

\subsection{Impact of rollout window size} \label{sec:results_ws}
To ensure a smooth transition from teacher-forced numerical inputs to autoregressive inputs during stochastic teacher forcing, training begins with a teacher forcing probability of $p = 1$, which is linearly decreased over time. We found this scheduling strategy to stabilize gradients and improve training dynamics. However, the choice of the rollout window size $S_\text{STF}$ used during stochastic teacher forcing was equally critical. 

\begin{figure}
\hspace{-10pt}\centerline{\includegraphics[width=0.58\columnwidth]{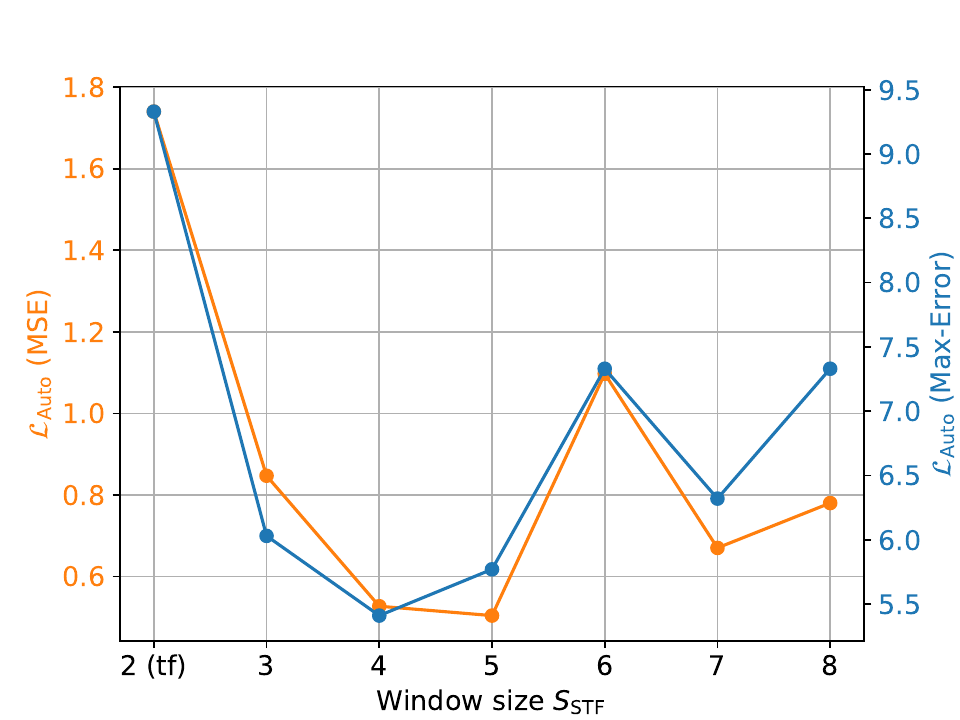}}
\caption{Autoregressive MSE and Maximum-Error over 10 test simulations (drawn from simulation scheme 2) as a function of the training-time rollout window size. For the first loss value (window size=2) the model was trained fully teacher-forced, for all other values the model was trained to minimize a stochastic teacher forcing loss $\mathcal{L}_{\text{STF}}$ for the respective rollout window size.}
\label{fig:results_ws}
\end{figure}

In Figure~\ref{fig:results_ws}, we report results from training surrogates with different rollout window sizes. All model and training hyperparameters were kept constant across runs, with the only variation being the window size $S_\text{STF}$ used for stochastic teacher forcing. The plot shows the resulting autoregressive test error $\mathcal{L}_{\text{Auto}}$ measured over 10 test simulations, as a function of the training-time window size. 

Intermediate rollout window sizes (4 and 5) yield the lowest test error, with window size 5 achieving the best autoregressive test MSE overall, i.e., $\mathcal{L}_\text{Auto}^\text{MSE} = 0.51$, suggesting an optimal trade-off between training stability and robustness to autoregressive inputs. In contrast, shorter windows (2–3), including fully teacher-forced training at window size 2, do not sufficiently expose the model to its own prediction errors, limiting its ability to learn effective correction strategies. Longer rollout windows ($\geq 6$), while potentially helpful to learn long-term dependencies, tend to destabilize gradients and training by introducing compounding errors too early, before the model has learned to generate reliable predictions. This was also reflected in higher maximum gradient norms observed for larger window sizes (e.g., about twice as large for window size 8 compared to 5).

\subsection{Hyperparameter-optimized surrogate} \label{sec:results_final}
The final surrogate model was trained using stochastic teacher forcing, employing the optimal rollout window size of $S_\text{STF} = 5$ as identified in the previous experiment, along with a linearly decaying teacher forcing probability.

\subsubsection{Performance on test set}
To assess the performance of this hyperparameter-optimized surrogate model, we conducted autoregressive rollouts on 40 simulations from the in-distribution test set. Across these rollouts, the autoregressive error metrics varied in the following ranges: MSE from $0.02$ to $0.31$, Hausdorff Distance from $0.97$~mm to $3.76$~mm, and maximum prediction error from $1.35$~mm to $4.73$~mm, where the maximum error was computed over all time steps, mesh nodes, and spatial directions. Averaged over all 40 rollouts, the final surrogate model achieved the following test metrics (mean ± standard deviation): $$ \mathcal{L}_{\text{Auto}}^{\text{MSE}} = 0.09 \pm 0.07, $$  $$ \mathcal{L}_{\text{Auto}}^{\text{Hausdorff}} = 1.81 \pm 0.52, $$  $$  \mathcal{L}_{\text{Auto}}^{\text{Max}} = 2.37 \pm 0.77.$$  

\begin{figure}
\centerline{\includegraphics[width=1\columnwidth]{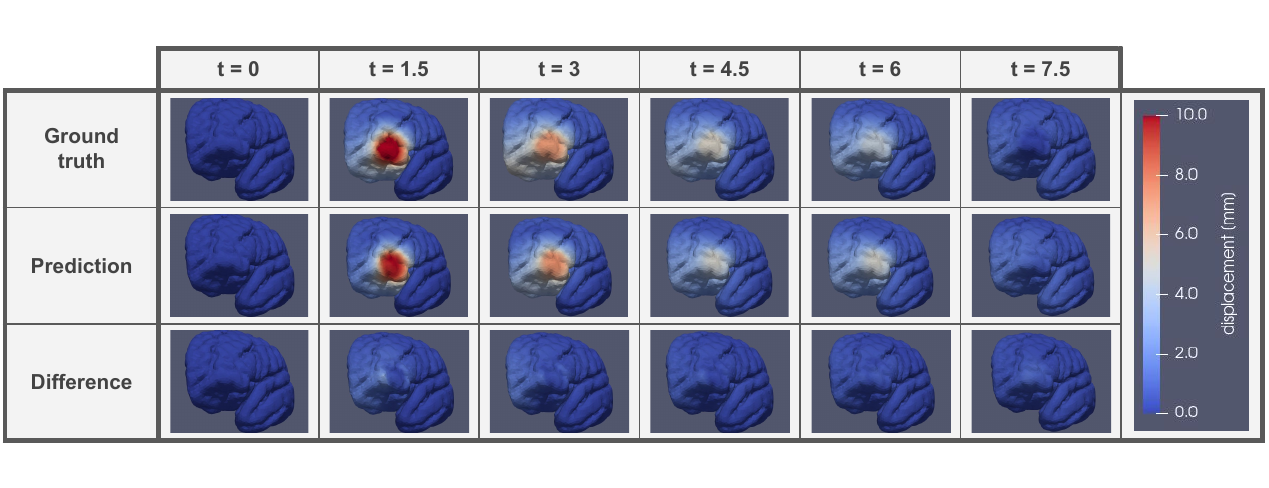}} \vspace{-10pt}
\caption{Example rollout on a test set deformation from simulation scheme 1, with an autoregressive MSE of $\mathcal{L}_{\text{Auto}}^{\text{MSE}} = \text{0.03}$.}
\label{fig:final_rollout}
\end{figure}

The model's inference time was approximately $18~\mathrm{ms}$ per step, supporting real-time model deployment with over 55 frames per second. Compared to the numerical simulations used for training, this corresponds to a speed up of over 90×, demonstrating the surrogate model’s significant computational efficiency. As a qualitative reference, Figure \ref{fig:final_rollout} shows an example rollout of the final surrogate on a test simulation, showcasing a high degree of visual similarity between the ground truth and the predicted brain deformations. Additional rollout examples from the test set are provided in Supplementary Figure S1 and Supplementary Figure~S2. Further, we also provide an exemplary rollout video as Supplementary Material.

\subsubsection{Generalization across schemes}
\begin{figure}
\centerline{\includegraphics[width=1\textwidth]{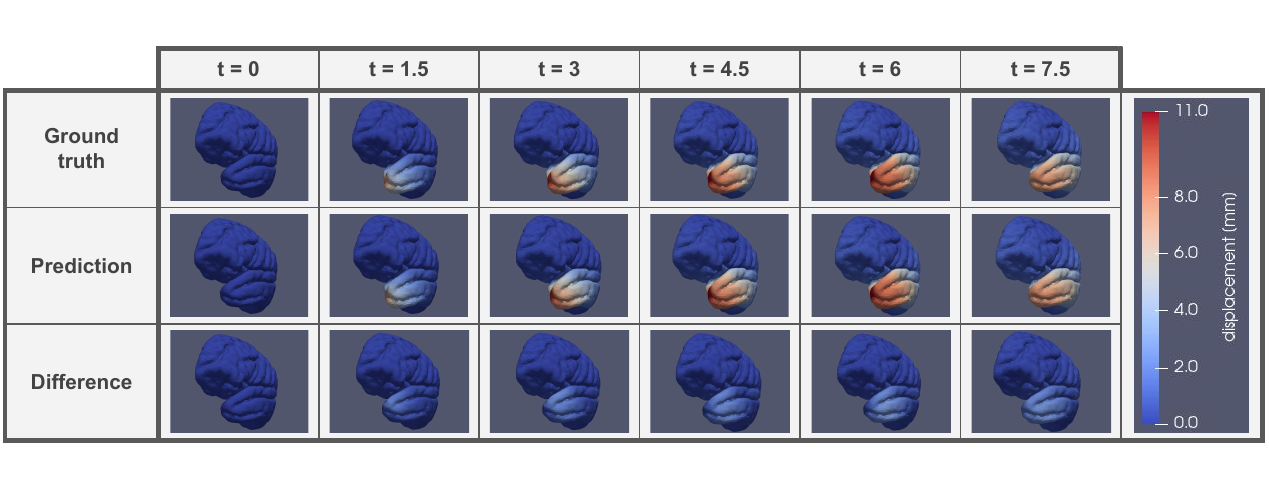}} \vspace{-10pt}
\caption{Example rollout on a cross-scheme test set deformation from simulation scheme 5, with an autoregressive MSE of $\mathcal{L}_{\text{Auto}}^{\text{MSE}} = \text{0.08}$.}
\label{fig:final_rollout_cross}
\end{figure}

The model’s ability to generalize across different instrument dynamics was evaluated on 10 test simulations from simulation scheme 5, which was not included in the training data. Averaging over all 10 autoregressive rollouts yielded:
$$ \mathcal{L}_{\text{Auto}}^{\text{MSE}} = 0.09 \pm 0.04,$$  $$ \mathcal{L}_{\text{Auto}}^{\text{Hausdorff}} = 1.99 \pm 0.27,$$ $$ \mathcal{L}_{\text{Auto}}^{\text{Max}} = 1.94 \pm 0.53.$$  These results are comparable to those observed on the in-distribution test set. As a qualitative example, Figure \ref{fig:final_rollout_cross} shows a rollout on a cross-scheme test simulation, with an MSE of $\mathcal{L}_{\text{Auto}}^{\text{MSE}} = 0.08$. An additional cross-scheme example is available in Supplementary Figure S3.

\subsubsection{Scalability with increased mesh complexity}
We evaluated the scalability of the surrogate model on a high-resolution brain mesh of the same subject, containing approximately 7 times more nodes ($K=151{,}201$). Using this refined mesh, we performed autoregressive rollouts with 10 simulations from scheme 1. Predictive accuracy remained high and comparable to the lower-resolution mesh:
$$ \mathcal{L}_{\text{Auto}}^{\text{MSE}} = 0.09 \pm 0.04, $$  $$ \mathcal{L}_{\text{Auto}}^{\text{Hausdorff}} = 2.03 \pm 0.30, $$  $$  \mathcal{L}_{\text{Auto}}^{\text{Max}} = 3.63 \pm 0.51.$$ An example rollout is shown in Supplementary Figure S4, illustrating strong visual agreement between the predicted and ground-truth dynamics. Despite the substantial increase in mesh size, the model remained computationally tractable for real-time use, with an inference time of 43~ms corresponding to 23~frames per second.

\subsubsection{Integration into interactive environment} \label{sec:results_integration}
\begin{figure}
\vspace{5pt}
\centerline{\includegraphics[width=0.48\columnwidth]{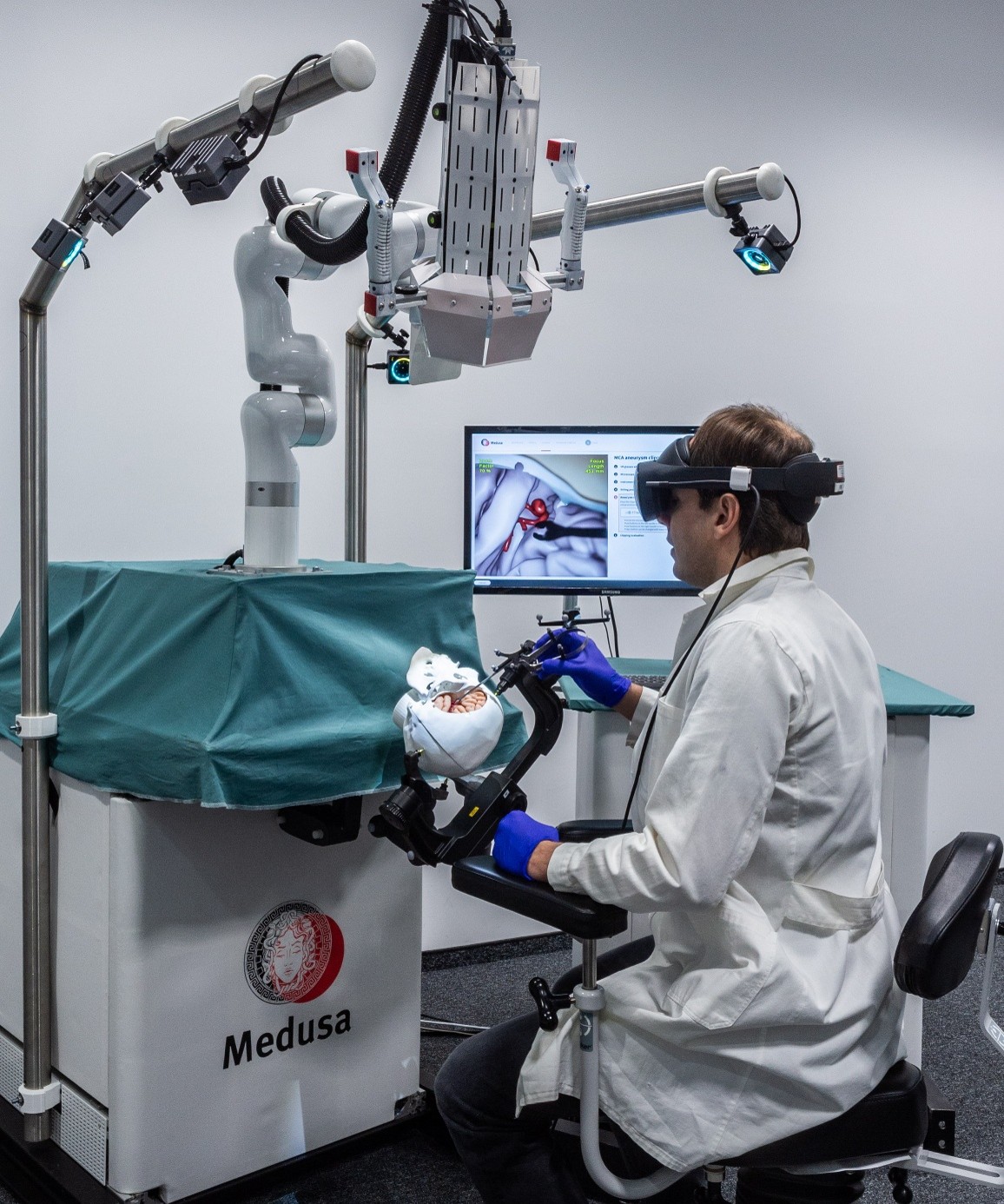}}
\caption{The neurosurgical training simulator~\cite{medusa} allows users to practice and refine complex clipping procedures of middle cerebral artery (MCA) aneurysms in a risk-free environment. The simulator consists of both a virtual simulation accessible through virtual reality glasses and a haptic interface. Deploying our trained surrogate model enables medical professionals to dynamically interact with the virtual brain and experience real-time deformations in response to their instrument movements (tracked via optical cameras).}
\label{fig:demonstrator}
\end{figure}

Finally, we successfully integrated the trained surrogate model into a neurosurgical training simulator environment, as shown in Figure~\ref{fig:demonstrator} and described in~\cite{medusa}. Since the simulator is built using Unity and C\texttt{++}, it was essential to ensure efficient deployment of the Python-trained surrogate model within this framework. To achieve this, we exported the PyTorch-based model using the Open Neural Network Exchange (ONNX) format~\cite{onnx} and developed a C\texttt{++} class that loads the model and performs inference based on the current configuration, i.e., the brain mesh displacement and the collision field resulting from interactions between surgical instruments and the brain.
Due to limitations in ONNX, the message accumulation step of the message passing module could not be included directly in the exported model. We addressed this by splitting the inference process into two stages and performing the message passing step in between using the torchscatter library~\cite{torchscatter}.

We evaluated the mean inference times on an NVIDIA RTX 3080 Ti GPU across various ONNX execution providers: TensorRT-RTX~\cite{TensorRT-RTX}, a dedicated backend for high-performance AI inference, delivered the best performance, with a mean inference time of 8~ms, which translates to a rendering speed of 125~frames per second. This was followed by CUDA (49~ms), TensorRT (59~ms), and CPU (414~ms). 

This preliminary integration enables users to interact dynamically with the virtual brain mesh using optically tracked instruments. Initial feedback from neurosurgeons testing the system has been highly positive, praising both the realism and responsiveness of the simulation.

\section{Discussion}
\label{sec:discussion}
In this work, we introduced an autoregressive deep learning-based surrogate model for real-time prediction of soft tissue dynamics as they occur within neurosurgical simulators. While prior work has explored deep learning-enhanced soft tissue deformation modeling~\cite{shahbazi_neural-augmented_2025,salehi_physgnn_2022,liu_real-time_2020,karami_real-time_2023,zhang_neural_2019}, our approach extends this line of research by employing a mesh-based yet highly scalable surrogate architecture and by explicitly addressing transient, dynamic simulation scenarios—--an aspect not previously demonstrated. 

We opted for UPTs as the backbone of our deep learning framework due to their architectural flexibility and scalability, particularly in handling large meshes where convolution-based methods often struggle to preserve geometric detail~\cite{Bronstein2021}. However, since transformer attention operations scale quadratically with the number of input tokens~\cite{Vaswani2017}, applying them directly at the full node level becomes computationally prohibitive for complex meshes with up to $151{,}000$ nodes. UPTs address this challenge by combining efficient graph-based feature aggregation at the supernode level, cross-attention-based down-projection, latent-space transformers, and a continuous field decoder. The supernode representation enables the efficient use of transformers at both the supernode and latent levels, which allow the model to capture long-range spatial dependencies, as already demonstrated in sequence modeling and computer vision~\cite{jumper2021, Dalla2023, Dosovitskiy2021}. Although UPTs allow incorporating additional information as model input (e.g., nodal velocity or acceleration), we found that this approach did not significantly improve model accuracy but only increased inference time. Overall, the described architectural design allows us to achieve inference times under 10~ms (a more than 200× speed-up over numerical simulations) without compromising accuracy or temporal stability. This substantial reduction in computation time is a key enabler for real-time applications, opening the door to fully interactive neurosurgical simulations. 

We generated a large and diverse dataset of transient deformation scenarios, closely reflecting realistic surgical behaviors. Specifically, we designed five medically motivated simulation schemes that capture a broad spectrum of brain deformation scenarios. These schemes varied not only in the location of instrument contact but also in the size of affected mesh regions, as well as in the direction of applied force. This variability was intended to expose the model to diverse instrument–tissue interaction patterns during training, as they can differ substantially across cases and subjects during inference. Our evaluation demonstrated strong generalization to unseen simulations (including an entirely withheld simulation scheme) highlighting the robustness of the model in capturing complex, previously unobserved tissue responses. Future work may further enhance generalizability by incorporating an even wider spectrum of deformation scenarios.

Switching from teacher-forced inputs during training to autoregressive inputs during inference causes a distribution shift, often leading to error accumulation, especially in longer rollouts. This challenge has been reported in various domains, including message passing surrogates for fluid simulations~\cite{brandstetter2023} and sequence modeling~\cite{bengio2015}. To address the distribution shift problem, we introduced \textit{stochastic teacher forcing}, a training strategy that gradually increases the reliance on the model’s own previous predictions. This helped the model adapt to its own output, significantly reducing error accumulation and enabling reliable multi-step rollouts. The ability to perform stable and accurate autoregressive predictions constitutes a key strength and innovation of our approach.

However, our findings also indicated that the naive inclusion of fully autoregressive inputs during training can destabilize gradient computation and hinder surrogate model optimization, due to error accumulation and the complexity of backpropagation through multiple time steps. To address this, we carefully tuned the rollout window size of the stochastic teacher forcing schedule. Our experiments showed that intermediate rollout window sizes of 4–5 steps provided the best balance between robustness to autoregressive inputs and training stability, outperforming both shorter and longer alternatives.

\subsection{Limitations}
We intentionally focused on a single brain geometry, selected to match the 3D-printed head in our neurosurgical simulator (see Figure \ref{fig:demonstrator}). While the surrogate model and its scalability were also assessed on a high-resolution mesh variant, the architectural flexibility of the deep learning framework is not confined to this geometry; re-training or fine-tuning for different subjects can be performed with minimal adaptation. As the primary objective of this proof-of-concept study was to develop and deploy an accurate, real-time-capable surrogate model for the neurosurgical simulator, the exploration of cross-subject generalizability is deferred to future work.

All instrument collision sites were sampled within a predefined region surrounding the Sylvian fissure. No interaction data was generated in other cortical areas. Although this limitation would impact the generalizability of the model for broader brain deformation tasks, it does not compromise the aim of our work, as the Sylvian fissure is of most importance during clippings of middle cerebral artery (MCA) aneurysms. Notably, the craniotomy in both the surgical simulator and the operating room is performed exclusively over this region of the scalp, further justifying our localized data sampling strategy. In future work, we plan to expand the simulation to include other parts of the brain beyond the MCA region, so it can support more types of neurosurgical procedures and become more flexible overall.

\section{Conclusion}
\label{sec:conclusion}
Surgical simulators have become an essential tool in modern medicine, significantly enhancing surgical training, preoperative planning, and overall procedural safety. This work addressed a critical requirement for such realistic simulator environments: the ability to simulate transient tissue-instrument interactions and the resulting deformation in real-time. Our proposed deep learning-based surrogate model effectively simulates brain tissue deformation dynamics autoregressively, achieving both high speed and accuracy. A central challenge was the transition from ground truth inputs during training to autoregressive inputs at inference time. To address this, we introduced a stochastic teacher forcing strategy that enables the surrogate model to better adapt to inference conditions and mitigates long-term error accumulation. Overall, our findings demonstrate that transformer-based surrogates offer both computational speed and simulation accuracy, paving the way for physically accurate and interactive surgical training environments.



\section*{Declaration of competing interests}
The authors declare that they have no known competing financial interests or personal relationships that could have appeared to influence the work reported in this paper.

\section*{Acknowledgments}
This project is financed by research subsidies granted by the government of Upper Austria within the research projects MIMAS.ai and MEDUSA (FFG grant no. 872604). RISC Software GmbH is a member of UAR (Upper Austrian Research) Innovation Network.


\bibliographystyle{unsrt}  
\bibliography{references}  

\end{document}